\documentclass[10pt,twocolumn,letterpaper]{article}

\usepackage{cvpr}
\usepackage{times}
\usepackage{epsfig}
\usepackage{graphicx}
\usepackage{amsmath}
\usepackage{amssymb}
\usepackage{array}
\usepackage{makecell}
\usepackage{dsfont}
\usepackage[pagebackref=true,breaklinks=true,letterpaper=true,colorlinks,bookmarks=false]{hyperref}

\cvprfinalcopy % *** Uncomment this line for the final submission

\def\cvprPaperID{1422} % *** Enter the CVPR Paper ID here

\ifcvprfinal\fi
\begin{document}

%%%%%%%%% TITLE
\title{Learning Local Image Descriptors with Deep Siamese and Triplet Convolutional Networks by Minimizing Global Loss Functions}

\author{Vijay Kumar B G, Gustavo Carneiro, Ian Reid\\
The University of Adelaide and Australian Centre for Robotic Vision\\
{\tt\small \{vijay.kumar,gustavo.carneiro,ian.reid\}@adelaide.edu.au}
}

\maketitle
%\thispagestyle{empty}

%%%%%%%%% ABSTRACT
\begin{abstract}

Recent innovations in training deep convolutional neural network (ConvNet) models have motivated the design of new methods to automatically learn local image descriptors.  The latest deep ConvNets proposed for this task consist of a siamese network that is trained by penalising misclassification of pairs of local image patches.
Current results from machine learning show that replacing this siamese by a triplet network can improve the classification accuracy in several problems, but this has yet to be demonstrated for local image descriptor learning.   
Moreover, current siamese and triplet networks have been trained with stochastic gradient descent that computes the gradient from individual pairs or triplets of local image patches, which can make them prone to overfitting. 
In this paper, we first propose the use of triplet networks for the problem of local image descriptor learning.  Furthermore, we also propose the use of a global loss that minimises the overall classification error in the training set, which can improve the generalisation capability of the model.
Using the UBC benchmark dataset for comparing local image descriptors, we show that the triplet network produces a more accurate embedding than the siamese network in terms of the UBC dataset errors.  Moreover, we also demonstrate that a combination of the triplet and global losses produces the best embedding in the field, using this triplet network.  Finally, we also show that the use of the central-surround siamese network trained with the global loss produces the best result of the field on the UBC dataset.
Pre-trained models are available online at \url{https://github.com/vijaykbg/deep-patchmatch}
\end{abstract}

%%%%%%%%% BODY TEXT
%------------------------------------------------------------------------
\section{Introduction}
The design of effective local image descriptors has been instrumental for the application of computer vision methods in several problems involving the matching of local image patches, such as wide baseline stereo~\cite{matas2004robust}, structure from motion~\cite{molton2004locally}, image classification~\cite{lowe1999object,sivic2003video}, just to name a few.  
Over the last decades, numerous hand-crafted~\cite{dalal2005histograms,lowe1999object,schmid1997local} and automatically learned~\cite{brownpami10,carneiro2010automatic,dosovitskiy2014discriminative,Han_cvpr15,masci2014descriptor,Simonyanpami14,trzcinski2013boosting,ZagoruykoCVPR15} local image descriptors have been proposed and used in the applications above.
Despite their conceptual differences, these two types of local descriptors are formed based on similar goals: descriptors extracted from local image patches of the same 3-D location of a scene must be unique (compared with descriptors from different 3-D locations) and robust to brightness and geometric deformations.
Given the difficulty in guaranteeing such goals for hand-crafted local descriptors~\cite{dalal2005histograms,lowe1999object,schmid1997local}, the field has gradually focused more on the automatic learning of such local descriptors, where an objective function that achieves the goals above is used in an optimisation procedure.  In particular, the most common objective function minimises the distance between the descriptors from the same 3-D location (\ie, same class) extracted under varying imaging conditions and different viewpoints and, at the same time, maximises that distance between patches from different 3-D locations (or different classes)~\cite{brownpami10,carneiro2010automatic,dosovitskiy2014discriminative,Han_cvpr15,masci2014descriptor,SimoSerra_2015ICCV,Simonyanpami14,trzcinski2013boosting,ZagoruykoCVPR15}. 

\begin{figure*}[t]
\begin{center}
\includegraphics[width=1.0\textwidth]{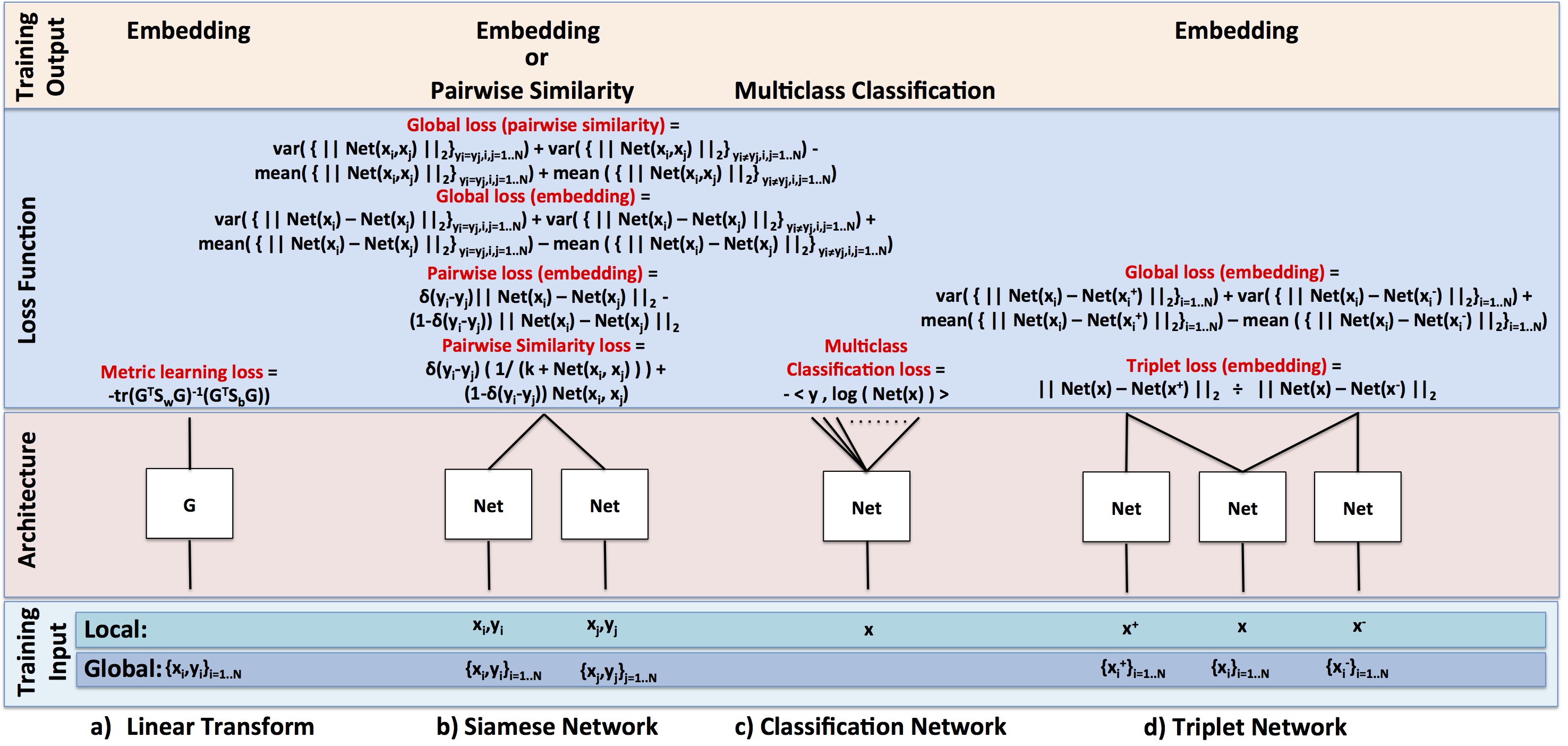} 
\end{center}
\caption{Comparison between different types of loss functions, network architectures and input/output types used by training methods of local image descriptor models.  The metric learning loss to learn a linear transform $\mathbf{G}$ is represented in (a)~\cite{brownpami10,carneiro2010automatic,Simonyanpami14,trzcinski2013boosting} (please see text for the definition of $\mathbf{S}_w$ and $\mathbf{S}_b$) and produces a feature embedding; in (b) we show the siamese network~\cite{Han_cvpr15,masci2014descriptor,ZagoruykoCVPR15} that can be trained with different loss functions and input types, where $\delta(.)$ denotes the Dirac delta function, $\text{y}$ is the data label, $\text{Net(x)}$ represents the ConvNet response for input $\text{x}$ (similarly for $\text{Net(x}_i\text{,x}_j\text{)}$), and the output can be an embedding (\ie, $\text{Net(x)}$) or a pairwise similarity estimation (\ie, $\text{Net(x}_i\text{,x}_j\text{)}$); the classification network in (c) can be used when classes of local image descriptors can be defined~\cite{dosovitskiy2014discriminative} and used in a multiclass classification problem; and in (d) the recently proposed triplet network~\cite{Wangcvpr14,hoffer2014deep,SchroffKPcvpr15,wohlhartcvpr15} is displayed with different loss functions and input types, where $\text{x}^+$ represents a point belonging to the same class as $\text{x}$ and $\text{x}^-$ a point from a different class of $\text{x}$ (this triplet net produces in general an embedding).  Note that our proposed global loss (embedding) in (b) and (d) takes the whole training set as input and minimises the variance of the distance of points belonging to the same and different classes and at the same time, minimise the mean distance of points belonging to the same class and maximise the mean distance of points belonging to different classes.  The global loss (pairwise similarity) in (b) is similarly defined (please see text for more details).}
\label{fig:intro}
\end{figure*}

The more recently proposed approaches~\cite{dosovitskiy2014discriminative,Han_cvpr15,masci2014descriptor,ZagoruykoCVPR15} based on deep ConvNets~\cite{krizhevsky2012imagenet} optimise slightly new objective functions that have the same goal as mentioned above.
Specifically, Zagoruyko and Komodakis~\cite{ZagoruykoCVPR15} and Han \etal~\cite{Han_cvpr15}  minimise a pairwise similarity loss of local image patches using a siamese network~\cite{bromley1993signature} (see Fig.~\ref{fig:intro}-(b)), where the patches can belong to the same or different classes (a class is for example a specific 3-D location).
Dosovitskiy \etal\cite{dosovitskiy2014discriminative} minimise a multi-class classification loss (Fig.~\ref{fig:intro}-(c)), where the model outputs the classification of a single input patch into one of the many descriptor classes (estimated in an unsupervised manner).
Moreover, Masci \etal~\cite{masci2014descriptor} propose a siamese network trained with a pairwise loss that minimises the distance (in the embedded space) between patches of the same class and maximises the distance between patches of different classes (Fig.~\ref{fig:intro}-(b)).
Even though these methods show substantial gains compared to the previous state of the art in public benchmark datasets~\cite{brownpami10,carneiro2010automatic,chen2005local,niyogi2004locality,Simonyanpami14,sugiyama2007dimensionality,trzcinski2013boosting}, we believe that the loss functions and network structures being explored for this task can be improved.  For instance,  the triplet network~\cite{Wangcvpr14,hoffer2014deep,SchroffKPcvpr15,wohlhartcvpr15} (see Fig.~\ref{fig:intro}-(d)) has been shown to improve  the siamese network on several classification problems, and the training of the siamese and triplet networks can involve loss functions based on global classification results, which has the potential to generalise better.

In this paper, we propose the use of the triplet network~\cite{Wangcvpr14,hoffer2014deep,SchroffKPcvpr15,wohlhartcvpr15} (Fig.~\ref{fig:intro}-(d)) and a new global loss function to train local image descriptor learning models that can be applied to the siamese and triplet networks (Fig.~\ref{fig:intro}-(b),(d)).  
The global loss to produce a feature embedding minimises the variance of the distance between descriptors (in the embedded space) belonging to the same and different classes, minimises the mean distance between descriptors belonging to the same class and maximises the mean distance between descriptors belonging to different classes (Fig.~\ref{fig:intro}-(b),(d)).  
For the case of pairwise similarity in siamese networks, this global loss minimises the variances of the pairwise similarity between descriptors belonging to the same and different classes, maximises the mean similarity between descriptors belonging to the same class and minimises the mean similarity between descriptors belonging to different classes (Fig.~\ref{fig:intro}-(b)).  
We first extend the siamese network~\cite{Han_cvpr15,masci2014descriptor,ZagoruykoCVPR15} to a triplet network, trained with a triplet loss~\cite{Wangcvpr14,hoffer2014deep,SchroffKPcvpr15,wohlhartcvpr15} and regularised by the proposed global loss (embedding).  Then we take the siamese network~\cite{Han_cvpr15,masci2014descriptor,ZagoruykoCVPR15} and train it exclusively with the global loss (pairwise similarity).  Finally, we take the central-surround siamese network~\cite{ZagoruykoCVPR15}, which is the current state-of-the-art model for the problem of local image descriptor learning, and train it with the global loss (pairwise similarity).
% Given that we use stochastic gradient descent (SGD) for learning the network, we approximate the global loss by computing the statistics of the mini-batch. 
We show on the public benchmark UBC dataset~\cite{UBC_dataset,brownpami10,snavely2008modeling} that: 1) the triplet network shows better classification results than the siamese network~\cite{bromley1993signature,masci2014descriptor,ZagoruykoCVPR15}; 2) the combination of the triplet and the global loss functions improves the results produced by the triplet loss from item (1) above, resulting in the best embedding result in the field for the UBC dataset; and 3) the global loss function used to train the central-surround siamese network~\cite{ZagoruykoCVPR15} produces the best classification result on the UBC dataset.

%------------------------------------------------------------------------

\section{Related Work}

In this section, we first discuss metric learning methods, which form the basis for several local image descriptor learning approaches.  Then, we discuss relevant local image descriptor learning methods recently proposed in the field, and highlight our contributions.

\subsection{Metric Learning}
\label{sec:metric_learning}

In general, metric learning (see Fig.~\ref{fig:intro}-(a)) assumes the existence of a set of points represented by $\{ \mathbf{x}_i \}_{i=1}^N$, with $\mathbf{x}_i \in \mathbb R^n$ and a respective set of labels $\{ y_i \}_{i=1}^N$, with $y_i \in \{1,...,C\}$, and the goal is to find a Mahalanobis distance with parameter $\mathbf{W}$.  For example, the square distance between $\mathbf{x}_i$ and $\mathbf{x}_j$ is~\cite{jain2012metric,weinberger2005distance}:
\begin{equation}
d_{\mathbf{W}} = (\mathbf{\mathbf{x}_i-\mathbf{x}_j})^\top \mathbf{W} (\mathbf{\mathbf{x}_i-\mathbf{x}_j}),
\label{eq:linear_DML}
\end{equation}
where the factorisation of the matrix $\mathbf{W} = \mathbf{G}\mathbf{G}^\top$ (with $\mathbf{G} \in \mathbb R^{n \times m}$) allows us to formulate the following optimisation problem: $\mathbf{G}^* = \arg \max_{\mathbf{G}} \text{tr} \left  ( (\mathbf{G}^\top \mathbf{S}_w \mathbf{G})^{-1} (\mathbf{G}^\top \mathbf{S}_b \mathbf{G}) \right ) $, with $ \mathbf{S}_k = \sum_{ij} \mathbf{Y}_k (\mathbf{\mathbf{x}_i-\mathbf{x}_j})(\mathbf{\mathbf{x}_i-\mathbf{x}_j})^\top$, 
$\mathbf{Y}_w = \mathbf{Y}$, $\mathbf{Y}_b = 1 - \mathbf{Y}$, and $\mathbf{Y}_{ij}=1$ if $y_i=y_j$ and $\mathbf{Y}_{ij}=0$, otherwise.  This optimisation is solved using the generalised Eigenvalue problem, which generates a linear feature transform that effectively produces a feature embedding.  
The method above has been extended in many ways, such that: 
1) it can handle multimodal distributions in~\cite{sugiyama2007dimensionality};
2) it optimises K nearest neighbour classification, which is formulated as a softmax loss minimisation and estimates a linear transform with eigenvalue decomposition~\cite{goldberger2004neighbourhood};
3) it optimises a large margin re-formulation of the problem in (\ref{eq:linear_DML}) using semidefinite programming~\cite{weinberger2005distance};  
4) it can use a prior for $\mathbf{W}$, which regularises the training and gets around the cubic complexity issues of the previous methods~\cite{davis2007information}; and 5) it can be extended to large problems using equivalence constraints~\cite{koestinger2012large}.
However, the main issue is the fact that (\ref{eq:linear_DML}) leads to a linear transformation that is unlikely to handle some of the difficult (and usually more interesting) learning problems.

Extending (\ref{eq:linear_DML}) to a non-linear transformation can be done by re-formulating $\mathbf{S}_k$ such that it involves inner products, which can then be kernelised~\cite{sugiyama2007dimensionality}, and the optimisation is again solved with generalised Eigenvalue problem~\cite{sugiyama2007dimensionality}.
Alternatively, this non-linear transform can be learned with a ConvNet using a siamese network~\cite{bromley1993signature} that minimises a pairwise loss~\cite{chopra2005learning} (Fig.~\ref{fig:intro}-(b)) by reducing the distance of patches (in the embedded space) belonging to the same class and increasing the distance of patches from different classes, similarly to the objective function derived from (\ref{eq:linear_DML}).  Note that this siamese network can produce either an embedding or a pairwise similarity estimation, depending on the architecture and loss function.  This siamese network has been extended to a triplet network that uses a triplet loss~\cite{Wangcvpr14,hoffer2014deep,SchroffKPcvpr15,wohlhartcvpr15} (Fig.~\ref{fig:intro}-(d)), which has been shown not only to produce the best classification results in several problems (e.g., STL10~\cite{coates2011analysis}, LineMOD~\cite{hinterstoisser2012gradient}, Labelled Faces in the Wild), but also to produce effective feature embeddings.

\subsection{Local Image Descriptor}

In the past, many local image descriptor learning methodologies have been proposed, with 
most based on the linear or non-linear distance metric learning, and explored in different ways~\cite{brownpami10,carneiro2010automatic,Simonyanpami14,trzcinski2013boosting}.  However, these methods have been shown to produce significantly worse classification results on the UBC dataset~\cite{UBC_dataset,brownpami10,snavely2008modeling} than the recently proposed siamese deep ConvNets~\cite{Han_cvpr15,masci2014descriptor,ZagoruykoCVPR15} (note that the UBC dataset is a benchmark dataset that has been used to compare local image descriptors).
Even though the triplet network~\cite{Wangcvpr14,hoffer2014deep,SchroffKPcvpr15,wohlhartcvpr15} has been demonstrated to improve the results produced by the siamese networks, it has yet to be applied to the problem of local image descriptor learning.
Finally, another relevant method is the discriminative unsupervised learning of local descriptors~\cite{dosovitskiy2014discriminative}, which uses a single deep ConvNet to classify input local patches into many classes, which are generated in an unsupervised manner (Fig.~\ref{fig:intro}-(c)).  However, the latter method has not been applied to the UBC dataset mentioned above.
It is also important to notice that none of the deep ConvNets methods above use the whole training set in a global loss function during the learning process, which can improve the generalisation ability of the model.

%------------------------------------------------------------------------
\section{Methodology}

As mentioned above in Sec.~\ref{sec:metric_learning}, we assume the existence of a training set of image patches and their respective classes, \ie, $\{ (\mathbf{x}_i,y_i) \}_{i=1}^N$, with $\mathbf{x}_i \in \mathbb R^n$ and $y_i \in \{1,...,C\}$ (note that we use $n$ as the patch size to simplify the notation, but the extension to a matrix representation for $\mathbf{x}$ is trivial).  The first goal of our work is to use a triplet network and respective triplet loss (defined below in detail)~\cite{Wangcvpr14,hoffer2014deep,SchroffKPcvpr15,wohlhartcvpr15} to produce a feature embedding $f(\mathbf{x},\theta_f)$ defined by $f: \mathbb R^n \times \mathbb R^k \rightarrow \mathbb R^m$, where $\theta_f \in \mathbb R^k$ denotes the network parameters (weights, biases and etc.).  The second goal is to design a new global loss function to train the triplet~\cite{Wangcvpr14,hoffer2014deep,SchroffKPcvpr15,wohlhartcvpr15} and siamese networks~\cite{Han_cvpr15,masci2014descriptor,ZagoruykoCVPR15}, where we are particularly interested in the 2-channel 2-stream network, represented by a multi-resolution central-surround siamese network.  Essentially, the siamese network can form a feature embedding, like the one above, or a pairwise similarity estimator, represented with $g(\mathbf{x}_i,\mathbf{x}_j , \theta_g)$, which is defined by $g: \mathbb R^n \times \mathbb R^n \times \mathbb R^k \rightarrow \mathbb R$.  In this section, we first explain the siamese and triplet networks, then we describe the proposed global loss function, and we also present the models being proposed in this paper.

\subsection{Siamese and Triplet Networks}
\label{sec:network_architecture}

The siamese network~\cite{bromley1993signature,Han_cvpr15,masci2014descriptor,ZagoruykoCVPR15} is trained with a two-tower deep ConvNet (Fig.~\ref{fig:intro}-(b)), where the weights on both towers are initialised at the same values and during stochastic gradient descent, they receive the same gradients (\ie, the weights on both towers are tied).  We consider the following definition of a deep ConvNet:
\begin{equation}
f(\mathbf{x},\theta_f) = f_{\text{out}} \circ r_{L} \circ h_L \circ f_L \circ ... \circ r_1 \circ h_1 \circ f_1(\mathbf{x}),
\label{eq:def_CNN}
\end{equation}
where the parameter $\theta_f$ is formed by the network weight matrices, bias vectors, and normalisation parameters for each layer $l \in \{1,...,L\}$, $f_l(.)$ represents the pre-activation function (\ie, the linear transforms in the convolutional layers), $h_l(.)$ represents a normalisation function, and $r_l(.)$ is a non-linear activation function (e.g., ReLU~\cite{nair2010rectified}).  Also note that $\mathbf{f}_l = [\mathbf{f}_{l,1},...,\mathbf{f}_{l,n_l}]$ is an array of $n_l$ pre-activation functions, representing the number of features in layer $l$.
A siamese network is then represented by two identical deep ConvNets trained using pairs of labelled inputs, where one possible loss function (called pairwise loss) minimises the distance between embedded features of the same class and maximises the distance between embedded features of different classes, as follows~\cite{chopra2005learning,masci2014descriptor}:
\begin{equation}
\begin{split}
J_1^s( \mathbf{x}_i,\mathbf{x}_j, & \theta_f ) = \delta(y_i-y_j)\| f^{(1)}(\mathbf{x}_i, \theta_f) - f^{(2)}(\mathbf{x}_j, \theta_f) \|_2   - \\
& (1 - \delta(y_i-y_j))\| f^{(1)}(\mathbf{x}_i, \theta_f) - f^{(2)}(\mathbf{x}_j, \theta_f) \|_2, 
\end{split}
\label{eq:loss_siamese_embedding}
\end{equation}
where $\delta(.)$ is the Dirac delta function and $f^{(1)}(\mathbf{x},\theta_f)$ is constrained to equal to $f^{(2)}(\mathbf{x},\theta_f)$.  Alternatively, the siamese network can be trained as a pairwise similarity estimator, with a pairwise similarity loss that can be defined as:
\begin{equation}
\begin{split}
J_2^s( \mathbf{x}_i,\mathbf{x}_j, \theta_g ) = & \delta(y_i-y_j)   (1/(\kappa+g(\mathbf{x}_i, \mathbf{x}_j, \theta_g ) ))   \\
&+(1 - \delta(y_i-y_j))g(\mathbf{x}_i, \mathbf{x}_j, \theta_g ) , 
\end{split}
\label{eq:loss_siamese_classifier}
\end{equation}
where the ConvNet $g(\mathbf{x}_i, \mathbf{x}_j, \theta_g)$ returns the similarity between the descriptors $\mathbf{x}_i$ and $\mathbf{x}_j$, with $\kappa$ representing a small positive constant.  
Note that the loss functions used by recently proposed methods~\cite{ZagoruykoCVPR15,Han_cvpr15} are conceptually similar to (\ref{eq:loss_siamese_classifier}), but not exactly the same, where the idea is to produce a ConvNet  $g(\mathbf{x}_i, \mathbf{x}_j, \theta_g)$ that returns large similarity values when the inputs belong to the same class and small values, otherwise. 
It is important to emphasise that the local descriptor learning model that currently produces the smallest error on the UBC dataset (Central-surround two-stream network) consists of a siamese network, trained with a loss similar to (\ref{eq:loss_siamese_classifier}), where the input patch is sampled twice at half the resolution of the input image: one sample containing the whole patch is input to the surround stream and another sample containing a sub-patch at the centre of the original patch is input to the central stream ~\cite{ZagoruykoCVPR15}. The output of these two streams are combined to obtain a similarity score.

The triplet network~\cite{Wangcvpr14,hoffer2014deep,SchroffKPcvpr15,wohlhartcvpr15} (Fig.~\ref{fig:intro}-(d)) is an extension of the siamese network that is trained with triplets at the input (which produces an embedding) using the triplet loss function, as follows:
\begin{equation}
\begin{split}
J_1^t  &  ( \mathbf{x},\mathbf{x}^+, \mathbf{x}^-, \theta_f ) =  \\ 
& \max \left ( 0,1-\frac{ \| f^{(1)}(\mathbf{x}, \theta_f) - f^{(3)}(\mathbf{x}^-, \theta_f) \|_2 } { \| f^{(1)}(\mathbf{x}, \theta_f) - f^{(2)}(\mathbf{x}^+, \theta_f) \|_2 + m }  \right ), \\
\end{split}
\label{eq:loss_triplet_embedding}
\end{equation}
where $m$ is the margin, $\mathbf{x}^+$ and $\mathbf{x}$ belong to the same class, $\mathbf{x}^-$ and $\mathbf{x}$ are from different classes, and $f^{(1)}(.)$, $f^{(2)}(.)$ and $f^{(3)}(.)$ are constrained to be the same network.  Note that the losses in (\ref{eq:loss_siamese_embedding}) and (\ref{eq:loss_triplet_embedding}) are apparently similar, but they have a noticeable difference, which is the fact that a triplet of similar and dissimilar inputs gives context for the optimisation process, as opposed to the pairwise loss that the siamese network minimises (same class) or maximises (different classes) as much as possible for each pair independently~\cite{hoffer2014deep}.

\subsection{Global Loss function}
\label{sec:global_loss_function}

The siamese and triplet networks presented in Sec.~\ref{sec:network_architecture} typically contain a large number of parameters, which means that a large number of pairs or triplets must be sampled from the training data such that a robust model can be learned.
However, sampling all possible pairs or triplets from the training dataset can quickly become intractable, 
where the majority of those samples may produce small costs in (\ref{eq:loss_siamese_embedding})-(\ref{eq:loss_triplet_embedding}), resulting in slow convergence~\cite{SchroffKPcvpr15}. 
An alternative is to have a smart sampling strategy, where one must be careful to avoid focusing too much on the hard training cases because of the possibility of overfitting~\cite{SchroffKPcvpr15,wohlhartcvpr15,Wangcvpr14}.
In this paper, we propose a simple, yet effective, loss function that can overcome these drawbacks.

\begin{figure}[t]
\begin{center}
\includegraphics[width=80mm]{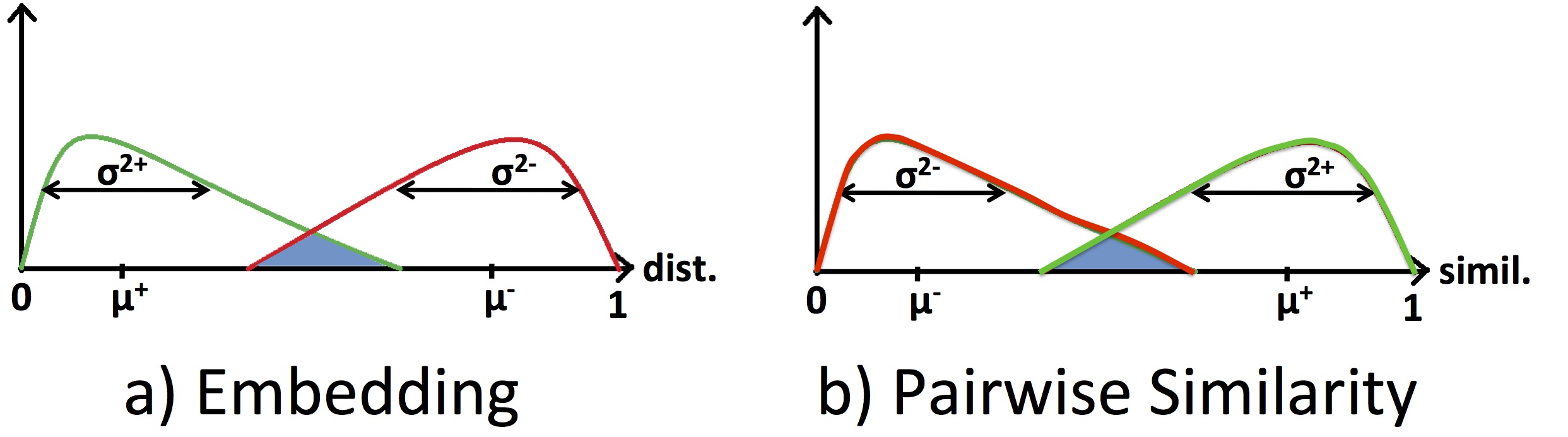} 
\end{center}
\caption{The objective of the proposed global loss is to reduce the proportion of false positive and false negative classification, which in the graph is represented by the area of the blue shaded region, 
assuming that the green curve indicates the distribution of distances in (a) or similarities in (b) between matching pairs (with mean $\mu^+$ and variance $\sigma^{2+}$) and the red curve denotes the distribution of non-matching pair distances in (a) and similarities in (b) (with mean $\mu^-$ and variance $\sigma^{2-}$).  Our proposed global loss for feature embedding (a) reduces the area mentioned above by minimising $\sigma^{2+}$, $\sigma^{2-}$ and $\mu^+$ and maximising $\mu^-$.  For the pairwise similarity in (b), the global loss minimises $\sigma^{2+}$, $\sigma^{2-}$ and $\mu^-$ and maximises $\mu^+$.}
\label{fig:global_loss}
\end{figure}

The main idea behind our proposed loss function is the avoidance of the over- or under-sampling problems mentioned above with the assumption that the distances (or similarities) between descriptors of the same class (i.e., matching pairs) or different classes (i.e., non-matching pairs) are samples from two distinct distributions. This allows us to formulate a loss function (for the embedding case) that globally tries to: 1) minimise the variance of the two distributions and the mean value of the distances between matching pairs, and 2) maximise the mean value of the distances between non-matching pairs.  Fig.~\ref{fig:global_loss}-(a) depicts the reasoning behind the design of the proposed global loss function, which is defined for the feature embedding case by:
\begin{equation}
\begin{split}
 J_1^g( \{ \mathbf{x}_i \}_{i=1}^N ,& \{ \mathbf{x}_i^+ \}_{i=1}^N, \{ \mathbf{x}_i^-\}_{i=1}^N , \theta_f ) = \\ &(\sigma^{2+}+\sigma^{2-})+\lambda \max\big(0,\mu^+ - \mu^- + t\big),
\end{split}
\label{eq:loss_global}
\end{equation}
where $\mu^+ = \sum_{i=1}^N d_i^+/N,~~ \mu^- = \sum_{i=1}^N d_i^-/N $, ~~
$\sigma^{2+} = \sum_{i=1}^N(d_i^+ -\mu^+)^2/N,~~~   \sigma^{2-} = \sum_{i=1}^N(d_i^- - \mu^-)^2/N$, with $\mu^+$ and $\sigma^{2+}$ denoting the mean and variance of the matching pair distance distribution,  
$\mu^-$ and $\sigma^{2-}$ representing the mean and variance of the non-matching pair distance distribution, 
 $d_i^+ = \frac{\lVert f^{(1)}(\mathbf{x}_i,\theta_f) - f^{(2)}(\mathbf{x}_i^+,\theta_f) \rVert_2^2}{4}$, $d_i^- = \frac{\lVert f^{(1)}(\mathbf{x}_i,\theta_f) - f^{(3)}(\mathbf{x}_i^-,\theta_f) \rVert_2^2}{4}$, 
$\lambda$ is a term that balances the importance of each term, $t$ is the margin between the mean of the matching and non-matching distance distributions and
$N$ is the size of the training set.
Note in (\ref{eq:loss_global}), that we assume a triplet network (i.e., $f^{(1)}(.)$, $f^{(2)}(.)$ and $f^{(3)}(.)$ are the same network), where the  
squared Euclidean distances of the matching and non-matching pairs of the $i^{th}$ triplet are constrained to be $0 \leq d_i^+, d_i^- \leq 1$ because of the division by 4, and the normalisation layer 
enforces that the norm of the embedding is 1.

Given that we use SGD for the optimisation process, we need to derive the gradient of the global loss function, as follows:
\begin{equation*}
\footnotesize{
\begin{split}
\frac {\partial  J_1^g}{\partial  f(\mathbf{x}_i)} &= -\frac{1}{2N}\Big[2(d_i^+ - \mu^+)\big(f(\mathbf{x}_i^+) - f(\mathbf{x}_i)\big)+ 2(d_i^- - \mu^-)\\
&\big(f(\mathbf{x}_i^-) - f(\mathbf{x}_i)\big) + \lambda (f(\mathbf{x}_i^+) - f(\mathbf{x}_i^-)) \mathds{1}((\mu^- - \mu^+) < t) \Big],\\
\end{split}}
\end{equation*}
\begin{equation}
%  \begin{split}
%   \frac {\partial  J_1^g}{\partial  f(\mathbf{x}_i^+)} = -\frac{1}{2N}\big[2\big((d_i^+ - \mu^+)&f(\mathbf{x}_i)\big) \\ 
% &+f(\mathbf{x}_i )\mathds{1}((\mu^- - \mu^+) < t) \big],\\ 
%  \frac {\partial  J_1^g}{\partial  f(\mathbf{x}_i^-)} = -\frac{1}{2N}\big[2\big((d_i^- - \mu^-)&f(\mathbf{x}_i)\big) \\ 
% &-f(\mathbf{x}_i) \mathds{1}((\mu^- - \mu^+) < t) \big]\\
%  \end{split}
\footnotesize{
\begin{split}
  \frac {\partial  J_1^g}{\partial  f(\mathbf{x}_i^+)} = -\frac{1}{2N}\Big[2(d_i^+ - &\mu^+)(f(\mathbf{x}_i) - f(\mathbf{x}_i^+)) \\
&+ \lambda (f(\mathbf{x}_i ) - f(\mathbf{x}_i^+))\mathds{1}((\mu^- - \mu^+) < t) \Big],\\ 
 \frac {\partial  J_1^g}{\partial  f(\mathbf{x}_i^-)} = -\frac{1}{2N}\Big[2(d_i^- - &\mu^-)(f(\mathbf{x}_i) - f(\mathbf{x}_i^-)) \\
&+ \lambda (f(\mathbf{x}_i^-) - f(\mathbf{x}_i))\mathds{1}((\mu^- - \mu^+) < t) \Big]\\
 \label{eq:loss_global_gradient}
 \end{split}}
\end{equation}
where 
the dependence on $\theta_f$ and the channel index $f^{(.)}$ are dropped to simplify the notation, and $\mathds{1}(a)$ is an indicator function with value $1$ when $a$ is true. It is important to note that the gradient 
${\partial  J_1^t}{/ \partial  f(\mathbf{x}_i)}$ of the triplet loss in (\ref{eq:loss_triplet_embedding}) depends only on the $i^{th}$
triplet of the training set, whereas the gradient ${\partial  J_1^g}{/ \partial  f(\mathbf{x}_i)} $ of the global loss in (\ref{eq:loss_global_gradient})
depends on $\mu^+$ and $\mu^-$, which in turn depends on the statistics of the samples in the whole training set. This dependence on global training set statistics has the potential to suppress the spurious gradients
computed from outliers and thus improving the generalisation of the trained model.

This global loss can be slightly modified to train a siamese network that estimates pairwise similarities, where the objective consists of: 1) minimising
the variance of the two distributions and the mean value of the similarities between non-matching pairs, and 2) maximising the mean value of the similarities between matching pairs.  Fig.~\ref{fig:global_loss}-(b) shows the idea behind the design of the proposed pairwise similarity global loss function, which is defined by:
\begin{equation}
\begin{split}
J_2^g = ( \{ \mathbf{x}_i \}_{i=1}^N ,& \{ \mathbf{x}_i^+ \}_{i=1}^N, \{ \mathbf{x}_i^-\}_{i=1}^N , \theta_f ) = \\ &(\sigma^{2+}+\sigma^{2-})+\lambda \max\big(0, m - (\mu^+ - \mu^-) \big),\\
\end{split}
\label{eq:loss_global_siamese}
\end{equation}
where $g(\mathbf{x},\tilde{\mathbf{x}},\theta_g)$ produces a similarity score between $\mathbf{x}$ and $\tilde{\mathbf{x}}$, $\mu^+ = \sum_{i=1}^N g(\mathbf{x}_i,\mathbf{x}_i^+,\theta_g)/N,~~ \mu^- = \sum_{i=1}^N g(\mathbf{x}_i,\mathbf{x}_i^-,\theta_g)/N $, ~~
$\sigma^{2+} = \sum_{i=1}^N(g(\mathbf{x}_i,\mathbf{x}_i^+) -\mu^+)^2/N,~~~   \sigma^{2-} = \sum_{i=1}^N(g(\mathbf{x}_i,\mathbf{x}_i^-) - \mu^-)^2/N$, with $\mu^+$ and $\sigma^{2+}$ denoting the mean and variance of the matching pair similarity distribution,  
$\mu^-$ and $\sigma^{2-}$ representing the mean and variance of the non-matching pair similarity distribution, 
$\lambda$ is a term that balances the importance of each term, $m$ is the margin between the mean of the matching and non-matching similarity distributions and
$N$ is again the size of the training set (note that we are abusing the notation with the re-definition of $\mu^+$, $\mu^-$, $\sigma^{2+}$, and $\sigma^{2-}$).  The gradient of this global loss function is derived as
\begin{equation}
 \begin{split}
  \frac{\partial  J_2^g}{\partial  g(\mathbf{x}_i,\mathbf{x}_i^+,\theta_g)} = \frac{2}{N}\big[ &
(g(\mathbf{x}_i,\mathbf{x}_i^+,\theta_g) -\mu^+)\\&-\frac{1}{2}\mathds{1}((\mu^+ -\mu^-) < m)\big]\\
  \frac{\partial  J_2^g}{\partial  g(\mathbf{x}_i,\mathbf{x}_i^-,\theta_g)} = \frac{2}{N}\big[& 
(g(\mathbf{x}_i,\mathbf{x}_i^-,\theta_g) -\mu^-)\\&+\frac{1}{2}\mathds{1}((\mu^+ -\mu^-) < m)\big].\\
 \end{split}
 \label{eq:loss_global_siamese_gradient}
\end{equation}

\subsection{Proposed Models}
\label{sec:proposed_models}

We propose {\bf four models} for the the problem of local image descriptor learning.  
The {\bf first model} consists of a triplet network trained with the triplet loss in (\ref{eq:loss_triplet_embedding}), which produces an embedding - this is labelled as {\bf TNet, TLoss}.
The {\bf second model} is a triplet network that also produces an embedding and uses the following loss function that combines the original triplet loss (\ref{eq:loss_triplet_embedding}) and the proposed global loss (\ref{eq:loss_global}) for the learning process:
\begin{equation}
\begin{split}
 &J_1^{tg}( \{ \mathbf{x}_i \}_{i=1}^N , \{ \mathbf{x}_i^+ \}_{i=1}^N, \{ \mathbf{x}_i^-\}_{i=1}^N ) = \\
 & \gamma \sum_j J_1^t ( \mathbf{x}_j,\mathbf{x}_j^+, \mathbf{x}_j^-)   +  J_1^g( \{ \mathbf{x}_i \}_{i=1}^N , \{ \mathbf{x}_i^+ \}_{i=1}^N, \{ \mathbf{x}_i^-\}_{i=1}^N ),\\
\end{split}
\label{eq:loss_combined}
\end{equation}
-- this model is labelled as {\bf TNet, TGLoss}.
The {\bf third model} is a siamese network that produces a similarity estimation of an input pair of local patches, but the model is trained with the siamese global loss defined in (\ref{eq:loss_global_siamese}) -- this model is labelled as {\bf SNet, GLoss}.  Finally, the {\bf fourth model} is the central-surround siamese network model described in Sec.~\ref{sec:network_architecture}  that also produces the pairwise similarity estimator of an input pair of local patches and is trained with the global loss (\ref{eq:loss_global_siamese}) -- this model is labelled as {\bf CS SNet, GLoss}.  
Note that for the first two models that produce the embedding, the comparison between local descriptors is done based on the $\ell_2$ norm of the distance in the embedded feature space.

In terms of the ConvNet structure, we use an architecture similar to the one described by Zagoruyko and Komodakis~\cite{ZagoruykoCVPR15}.  Specifically, the triplet network has the following structure:
B(96,7,3)-P(2,2)-B(192,5,1)-P(2,2)-B(256,3,1)-B(256,1,1)-B(256,1,1). The siamese network has the following architecture: B(96,7,3)-P(2,2)-B(192,5,1)-P(2,2)-B(256,3,1)-B(256,1,1)-C(1,1,1). Furthermore, the central-surround siamese network has the following structure: B(95,5,1)-P(2,2)-B(96,3,1)-P(2,2)-B(192,3,1)-B(192,3,1) and the final block that combines the outputs 
of the two input streams has components B(768,2,1)-C(1,1,1). In the description above, P($p,q$) is a max pooling layer of size $p \times p$ and stride $q$, and B($n,k,s$) is a block with the components C($n,k,s$)-bnorm($n$), where C($n,k,s$)
is a convolutional layer with $n$ filters of kernel size $k$ and stride $s$, bnorm($n$) is the batch normalisation unit \cite{IoffeSicml15} with $2n$ parameters. Each $B$ and $C$ is followed by a rectified
linear unit~\cite{nair2010rectified} except the final layer. Finally, the output feature from the exmedding networks are normalised to have unit norm, as mentioned in Sec.~\ref{sec:global_loss_function}.

\section{Toy Problem}
\label{sec:toy_problem}

\begin{figure*}[ht]
 \centering
\begin{tabular}{cccc}
\hspace{-6mm}
\includegraphics[trim = 0.5in 0.3in 0.5in 0.1in, clip,width=30mm]{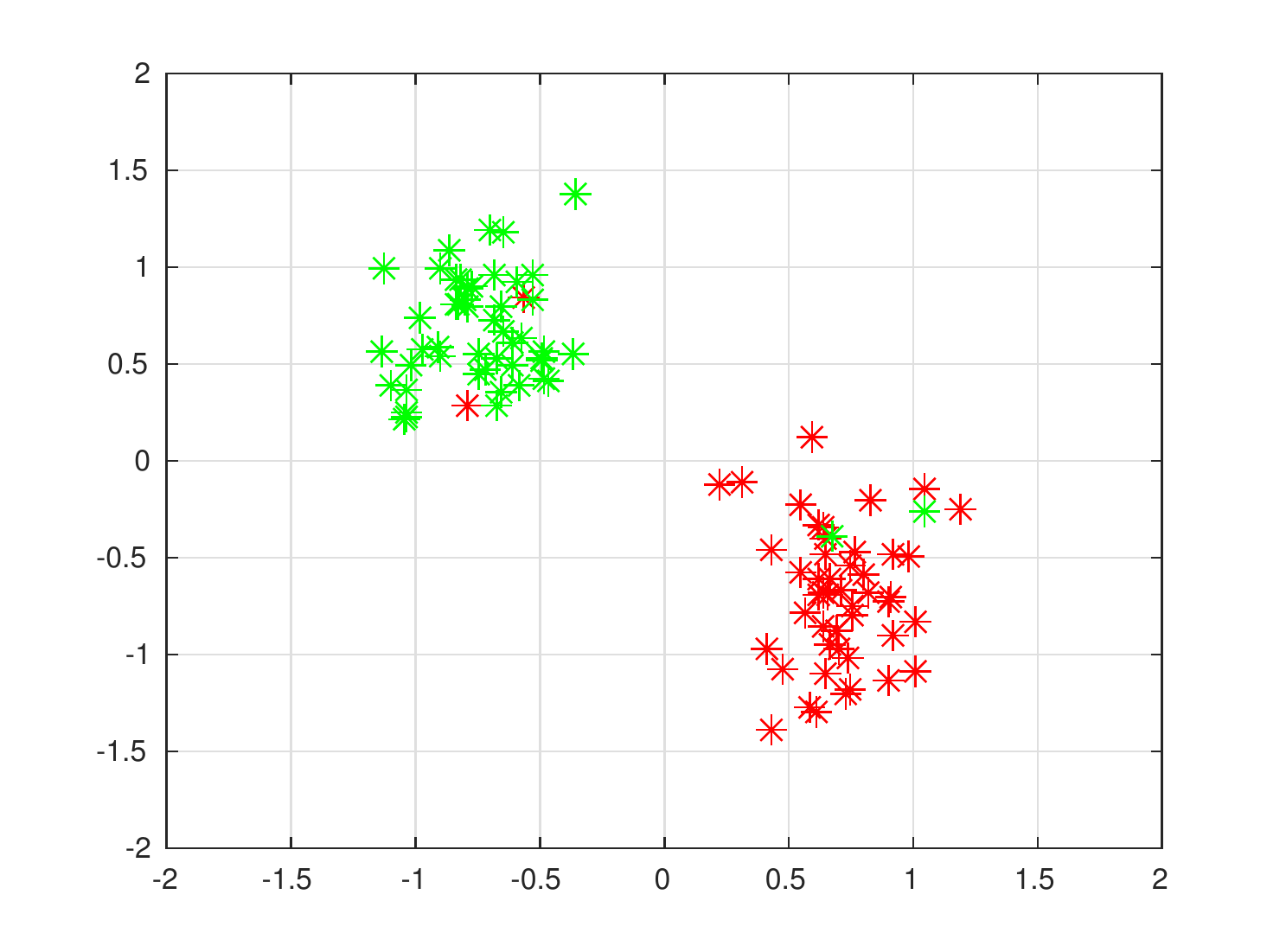}&
\includegraphics[trim = 0.5in 0.3in 0.5in 0.1in, clip,width=30mm]{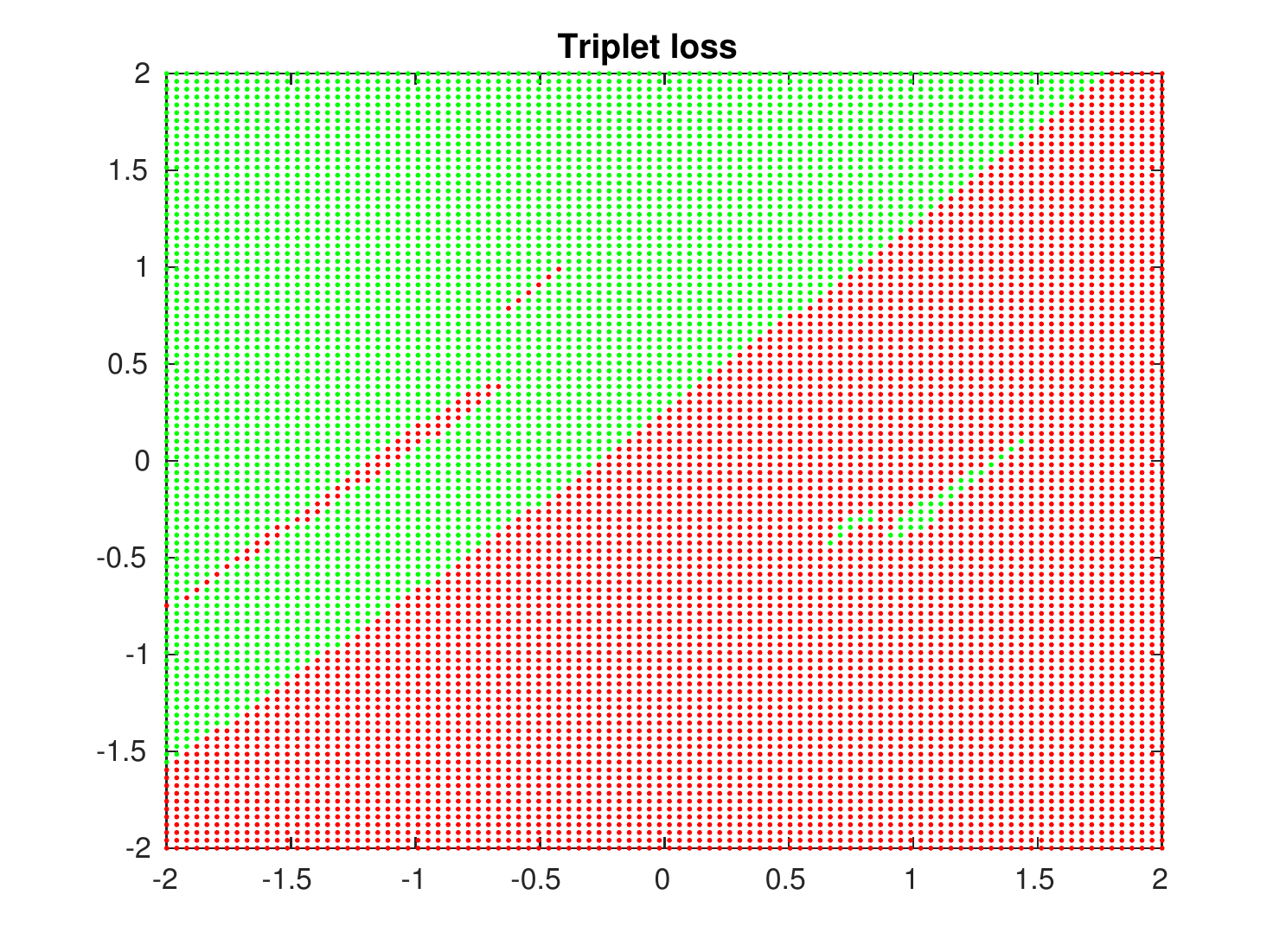}&
\includegraphics[trim = 0.5in 0.3in 0.5in 0.1in, clip,width=30mm]{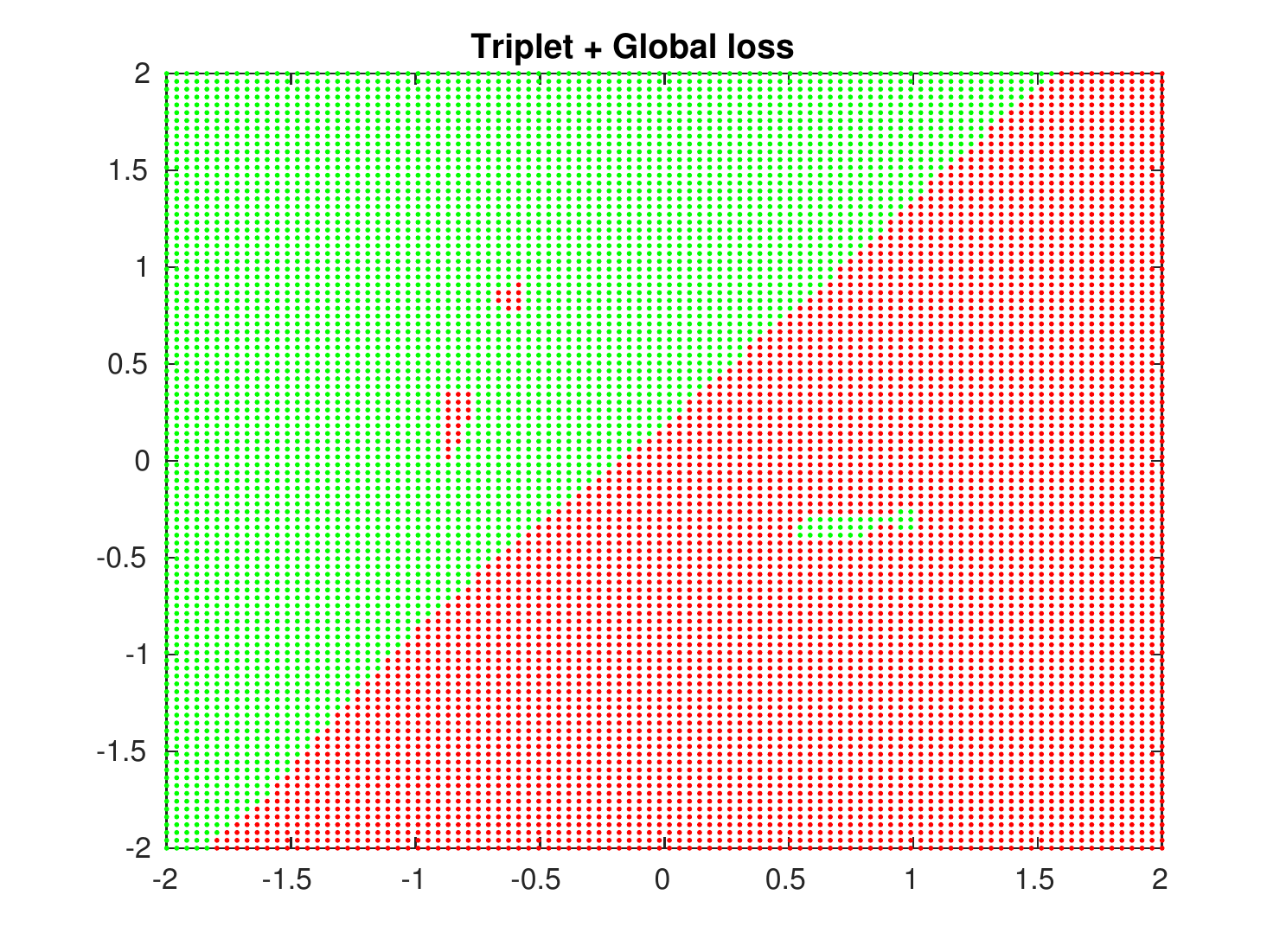}&
\includegraphics[trim = 0.5in 0.3in 0.5in 0.1in, clip,width=30mm]{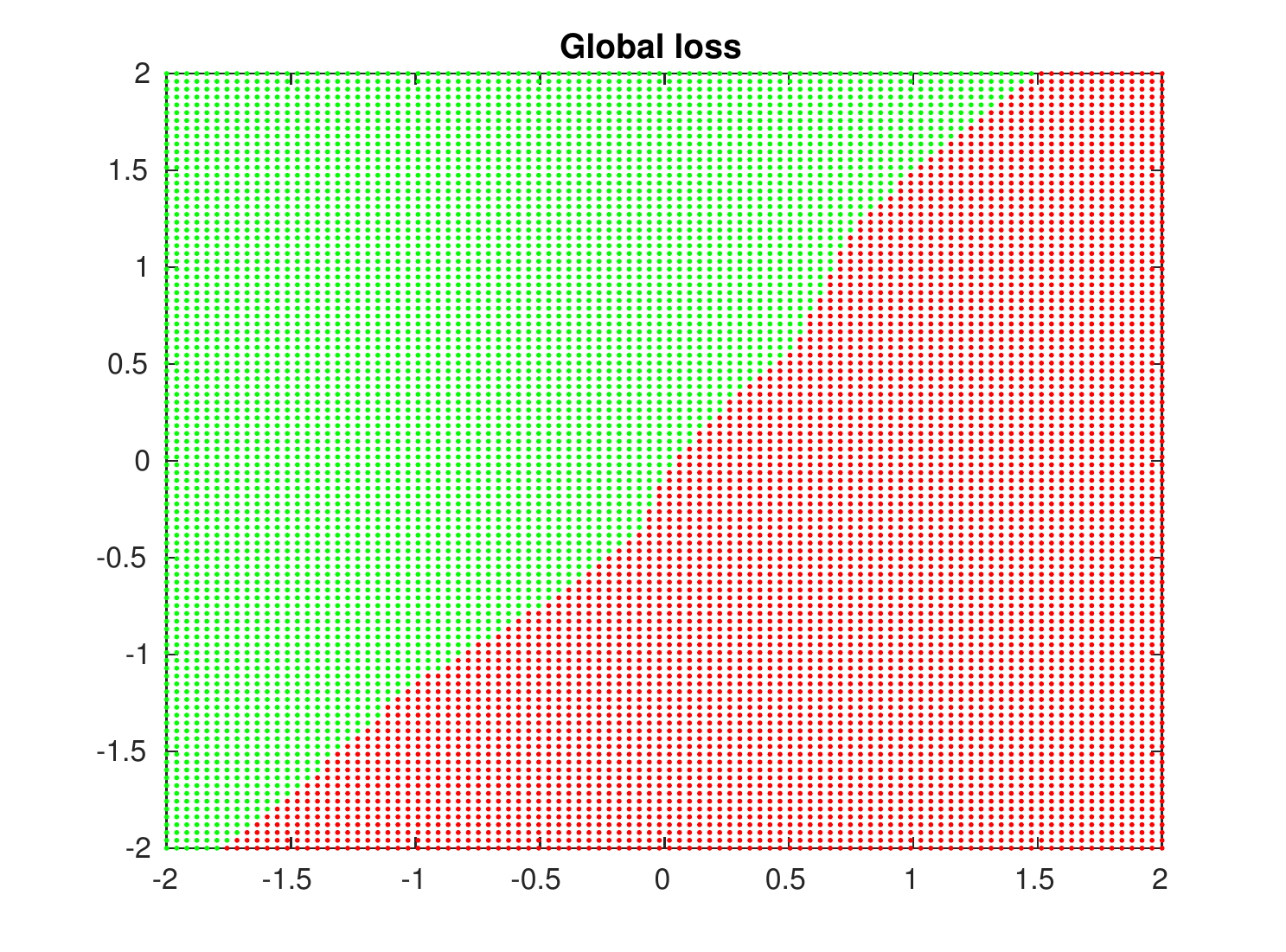}\\
a) Training set & b) Triplet loss (\ref{eq:loss_triplet_embedding}) & c) Combined triplet and global losses (\ref{eq:loss_combined}) & d) Global loss (\ref{eq:loss_global})\\
\end{tabular}
\caption{Illustration to compare the robustness of different loss functions to outliers: (a) training data with outliers, (b-d) classification of points in the input space based on the nearest neighbour classifier run
in the embedding space learned with the triplet loss (b), the combined triplet and global losses (c), and the global loss (d).}
\label{fig_toy}
\end{figure*}

To illustrate the robustness of the proposed global loss function to outliers, we generated a toy dataset in two dimensions with two classes (80 samples from two Gaussian distributions) represented by two distinct cloud of points, as indicated by the red and green points in Fig. \ref{fig_toy}-(a). We introduce outliers
by switching the labels of randomly selected points (\ie, we switch the labels of $5\%$ of the training set, or $4$ samples). We generate a set of triplets from this training set and train a ConvNet that maps the input points to an output embedding space with $128$ dimensions with the following structure: B(256,2,1)-B(512,1,1)-C(128,1,1), where the output is normalised to have unit norm and these blocks are defined in Sec.~\ref{sec:proposed_models}.
Three separate trainings are run: the first training uses the triplet loss function in (\ref{eq:loss_triplet_embedding}), the second uses a combination of the triplet and global losses in (\ref{eq:loss_combined}), and the third uses only the global loss in (\ref{eq:loss_global}). To ensure a fair comparison, we run the experiments with identical settings, where the only difference is the loss function.
We evaluate the models learned from each loss function by computing the embedding of a grid of points from the input space, and labelling them based on the label of the nearest neighbour from the training set, found in the embedding space.

Figure~\ref{fig_toy}-(b) shows the input space labelled according to the nearest neighbour classifier run in the embedding space generated by the triplet loss.  Similarly, Fig.~\ref{fig_toy}-(c) shows the same result for the combined triplet and global losses and Fig.~\ref{fig_toy}-(d) displays the results for the global loss.
In general, it is clear that outliers affect more the classifier in (b), which seems to be over-fitting the training data.
Such labelling mistakes are reduced when we use the combination of the triplet and global losses as show in Fig.~\ref{fig_toy}-(c). The label map in Fig.~\ref{fig_toy}-(d) generated by the embedding that uses global loss is coherent
even at the locations, where outliers can be found in the training set, indicating that the global loss function is robust to outliers.

\begin{table*}[ht!]
\begin{center}
 \begin{tabular}{|m{1.3cm}| m{1.3cm}| m{1cm}| m{0.8cm}| m{0.8cm}| m{0.6cm}|m{0.7cm}|m{0.8cm}|m{0.8cm}|m{0.7cm}|} 
 \hline
 \multicolumn{2}{|c|}{Datasets} & \multicolumn{2}{c|}{\small{Proposed Models}} & \multicolumn{4}{c|}{\small{Zagoruyko} \etal \cite{ZagoruykoCVPR15}} & \multicolumn{2}{c|}{\small{Xufeng} \etal \cite{Han_cvpr15}}\\
  \hline
  Train & Test & \small{CS~SNet, GLoss}& \small{SNet, GLoss}  & \small{2ch-2stream }& \small{2-ch} & \small{Siam} & \small{siam-2stream} & \small{4096d-F(512)} &\small{512d-F(512)} \\
 \hline   
 \small{Liberty} & \small{Notredame} &$\mathbf{0.77}$& $1.84$   &    $1.9$       &  $3.03$        & $4.33$     & $3.05$               & $3.87$   &  $4.75$    \\
 \hline
 \small{Liberty} & \small{Yosemite} &$\mathbf{3.09}$&   $6.61$       &       $5.00$       & $7$           &   $14.89$     & $9.02$       &  $10.88$ & $13.58$     \\
 \hline
 \small{Notredame}& \small{Liberty }&$\mathbf{3.69}$&    $6.39$      &     $4.85$      &$6.05$         &   $8.77$      &$6.45$        &   $6.9$  & $8.84$          \\
 \hline
 \small{Notredame}&\small{Yosemite} &$\mathbf{2.67}$&    $5.57$    &     $4.10$       &$6.04$          &   $13.23$      & $10.44$        &  $8.39$   & $11$      \\
 \hline
 \small{Yosemite} & \small{Liberty} &$\mathbf{4.91}$&    $8.43$      &        $7.2$      &    $8.59$     &  $13.48$       & $11.51$     &  $10.77$  & $13.02$       \\
 \hline
 \small{Yosemite}&\small{Notredame} &$\mathbf{1.14}$&   $2.83$       &    $2.11$       &    $3.05$      &   $5.75$      & $5.29$       &  $5.67$    & $7.7$            \\
 \Xhline{2\arrayrulewidth}
 \multicolumn{2}{|c|}{\textbf{mean}}   &  $\mathbf{2.71}$ &  $5.28$      &   $4.19$      &  $5.63$          & $10.07$     & $7.63$     &  $7.75$     &   $9.82$   \\
\hline
 \end{tabular}
 \bigskip
 \caption{{\bf Pairwise similarity results:} False Positive Rate at 95\% recall (FPR95) on UBC benchmark dataset, where bold numbers indicate the best results on the dataset. Note that for our models, we use the test set specified in~\cite{UBC_dataset} to compute these values.}
 \label{tbl:table_fpr_pairwise}
\end{center}
\end{table*}

\begin{table}[ht!]
\begin{center}
\tabcolsep=0.05cm
 \begin{tabular}{|c| c| m{1cm}| m{1cm}|m{1cm}|m{1cm}|m{1cm}|} 
 \hline
 \multicolumn{2}{|c|}{\scriptsize{Datasets}} & \multicolumn{2}{c|}{\scriptsize{Proposed Models}} & \multicolumn{2}{c|}{\scriptsize{Zagoruyko} {et.al.}\cite{ZagoruykoCVPR15}} & \scriptsize{Simonyan {et.al.}} \cite{Simonyanpami14}\\
  \hline
  \scriptsize{Train} & \scriptsize{Test} & \scriptsize{TNet, TGLoss} & \scriptsize{TNet, TLoss} & \scriptsize{siam-2stream-l2} & \scriptsize{siam-l2} &\scriptsize{discr. proj.}\\
 \hline   
 \scriptsize{Liberty} & \scriptsize{Notredame} &  \scriptsize{$\mathbf{3.91}$ }    &   \scriptsize{$4.47$ } & \scriptsize{$4.54$ }  &  \scriptsize{$6.01$}    &\scriptsize{$7.22$} \\
 \hline
 \scriptsize{Liberty} & \scriptsize{Yosemite} &     \scriptsize{$\mathbf{10.65}$ } &     \scriptsize{$11.82$ } & \scriptsize{ $13.24$ } &      \scriptsize{$19.91$}    & \scriptsize{$11.18$ }    \\
 \hline
 \scriptsize{Notredame}& \scriptsize{Liberty } &    \scriptsize{   $9.91$ }   &     \scriptsize{ $10.77$ }   &    \scriptsize{ $\mathbf{8.79}$} &   \scriptsize{ $13.24$}   &    \scriptsize{$12.42$}      \\
 \hline
 \scriptsize{Notredame}&\scriptsize{Yosemite} &   \scriptsize{$\mathbf{9.47}$ }   &     \scriptsize{  $10.96$}    &     \scriptsize{ $13.02$}   & \scriptsize{ $12.64$ }  & \scriptsize{$10.08$  }    \\
 \hline
 \scriptsize{Yosemite} & \scriptsize{Liberty} &   \scriptsize{   $13.45$}   &      \scriptsize{ $13.9$ }    &  \scriptsize{$\mathbf{12.84}$ } &  \scriptsize{ $17.25$}    &  \scriptsize{$14.58$ }     \\
 \hline
 \scriptsize{Yosemite}&\scriptsize{Notredame} &   \scriptsize{ $\mathbf{5.43}$}   &     \scriptsize{   $5.85$}   &   \scriptsize{ $5.58$ }  &   \scriptsize{ $ 8.38$  }  & \scriptsize{$6.82$ }      \\
 \Xhline{2\arrayrulewidth}
 \multicolumn{2}{|c|}{\scriptsize{\textbf{mean}}}   &    \scriptsize{  $\mathbf{8.8}$}   &    \scriptsize{ $9.63$}  &   \scriptsize{  $9.67$ } &   \scriptsize{ $13.45$ }   & \scriptsize{$10.38$ }    \\
\hline
 \end{tabular}
 \bigskip
 \caption{{\bf Embedding results:} False Positive Rate at 95\% recall (FPR95) on UBC benchmark dataset, where bold numbers indicate the best results on the dataset.  Note that for our models, we use the test set specified in~\cite{UBC_dataset} to compute these values.}
 \label{tbl:table_fpr_embedding}
\end{center}
\end{table}

\section{Experiments}

In this section, we first describe the dataset used for assessing our proposed models, then we explain the model setup, followed by a presentation of the results.

\subsection{UBC Benchmark Dataset}

The experiments are based on the performance evaluation of local image patches using the standard UBC benchmark dataset~\cite{UBC_dataset,brownpami10,snavely2008modeling}, which contains three sets: Yosemite, Notre Dame, and Liberty.  Using these sets, we run six combinations of training and testing sets, where we use one set for training and another for testing.  Each one of this sets has more than $450,000$ local image patches (with normalised orientation and scale) of size $64 \times 64$ sampled using a Difference of Gaussians (DoG) detector.  In each of these sets there are more than $100,000$ patch classes that are determined based on their 3-D locations obtained from multi-view stereo depth maps.  These patch classes are used to produce $500,000$ pairs of matching (\ie, from the same class) and non-matching (\ie, different classes) image patches.  Each model is assessed using the false positive at $95\%$ recall (FPR95) on each of the six combinations of training and testing sets, the mean over all combinations, and the receiver operating characteristic (ROC) curve also for each of the six combinations. The test set contains $100,000$ pairs with equal number of matching and non-matching pairs and is chosen as specified in~\cite{UBC_dataset}.

\subsection{Training Setup and Implementation Details}
\label{subsec:experiments}

The model training is based on stochastic gradient descent (SGD) that involves: 1) the use of a learning rate of $0.01$ that gradually (automatically computed based on the number of epochs set for training) decreases after each epoch until it reaches $0.0001$; 2) a momentum set at $0.9$, 3) weight decay of $0.0005$, and 4) data augmentation by rotating the pair of patches by $90$, $180$, and $270$ degrees, and flipping the images horizontally and vertically (\ie, augmented 5 times: 3 rotations and 2 flippings)~\cite{ZagoruykoCVPR15}.
The training set for the triplet and siamese networks consists of a set of $250,000$ triplets, which are  sampled randomly from the aforementioned set of $500,000$ pairs of matching and non-matching image patches, where it is important to make sure that the triplet contains one pair of matching image patches and one patch that belongs to a different class of this pair.
The mini-batch of the SGD optimisation consists of $250$ triplets (randomly picked from this $250K$ set of triplets), which is used to compute the global loss in (\ref{eq:loss_global})~and~(\ref{eq:loss_global_siamese}).
Our Matlab implementation takes $\approx56$ hours for training a model and processes $16K$ images/sec during testing on a GTX 980 GPU.

The triplet networks {\bf TNet-TLoss} and {\bf TNet-TGLoss} use the three towers of ConvNets (see Fig.~\ref{fig:intro}) to learn an embedding of size 256 (we choose this number of dimensions based on the feature dimensionality of the models in~\cite{ZagoruykoCVPR15}, which also have $256$ dimensions before the fully connected layer).  During testing, only one tower is used (all three towers are in fact the same after training) to compute the embedded features, which are compared based on the $\ell_2$ norm of the distance between these embedded features.
The network weights for the {TNet-TLoss} network are initialised randomly and trained for 100 epochs, whereas the weights for the {TNet-TGLoss} network are trained for 50 epochs after being initialised using the weights from {TNet-TLoss} network trained for 50 epochs (the initialisation from the {TNet-TLoss} model trained with early stopping provided a good initialisation for {TNet-TGLoss}).
 This number of epochs for training is decided based on the convergence obtained in the training set with respect to the loss function.
Moreover, the margin parameter $m=0.01$ in (\ref{eq:loss_triplet_embedding}) and $\gamma=1$, $t=0.4$ and $\lambda=0.8$ in (\ref{eq:loss_combined}) are estimated via cross validation.
For the siamese networks {\bf SNet-GLoss} and {\bf CS-SNet-GLoss}, the weights are randomly initialised and trained for 80 epochs (again, based on the convergence of the training set).  Finally, $m=1$ and 
$\lambda=1$ in (\ref{eq:loss_global_siamese}) are also estimated via cross validation.

\begin{figure*}
\begin{center}
\begin{tabular}{ccc}
\includegraphics[width=0.27\textwidth]{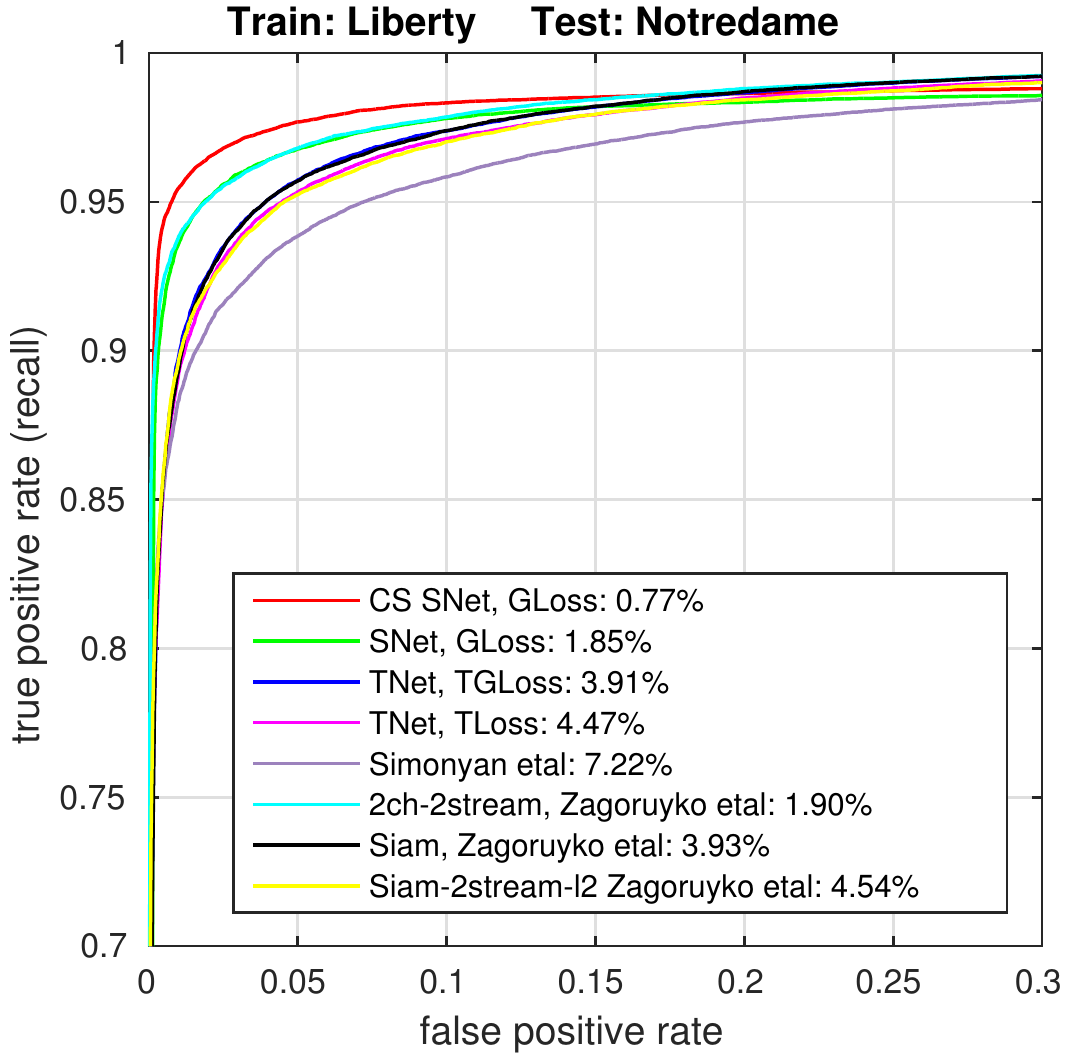} &
\includegraphics[width=0.27\textwidth]{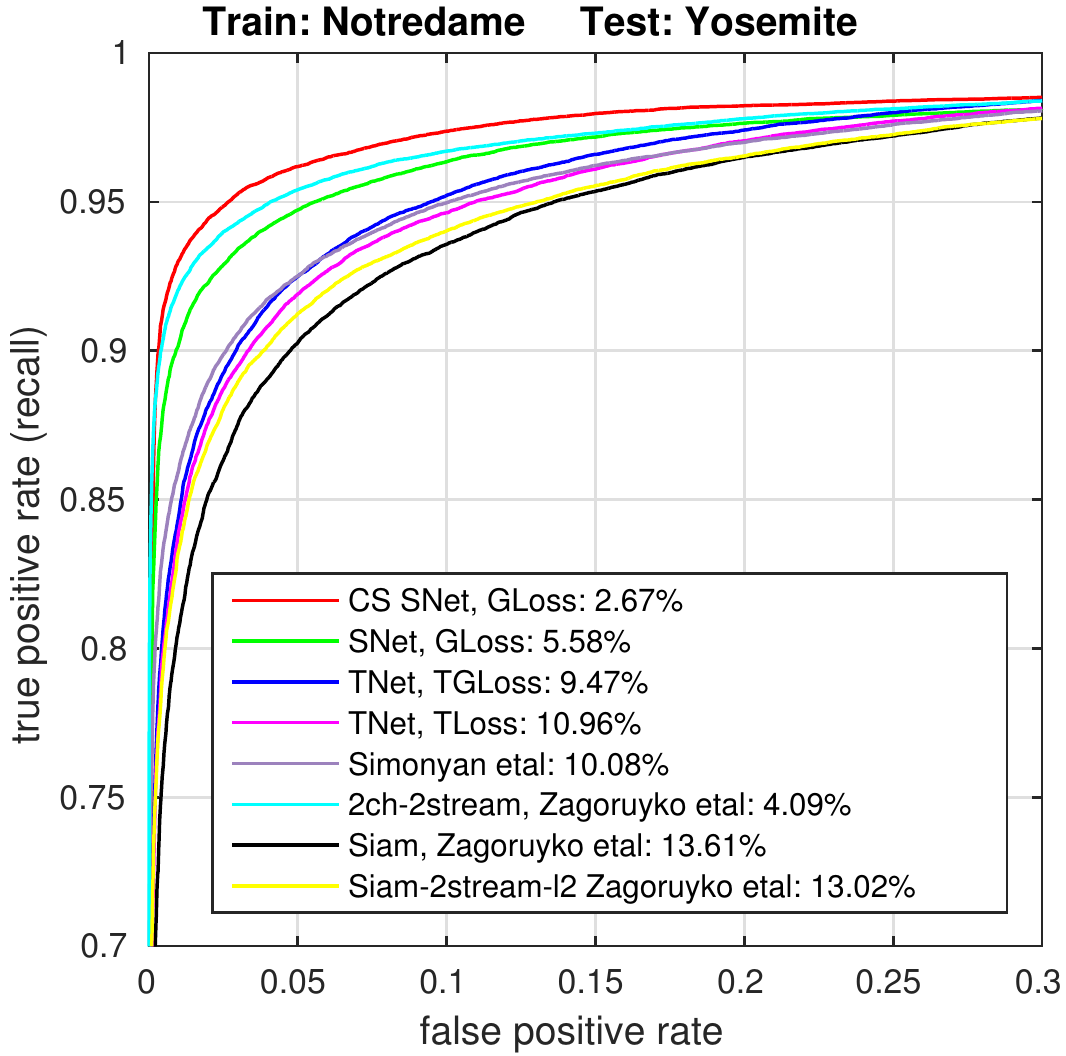} &
\includegraphics[width=0.27\textwidth]{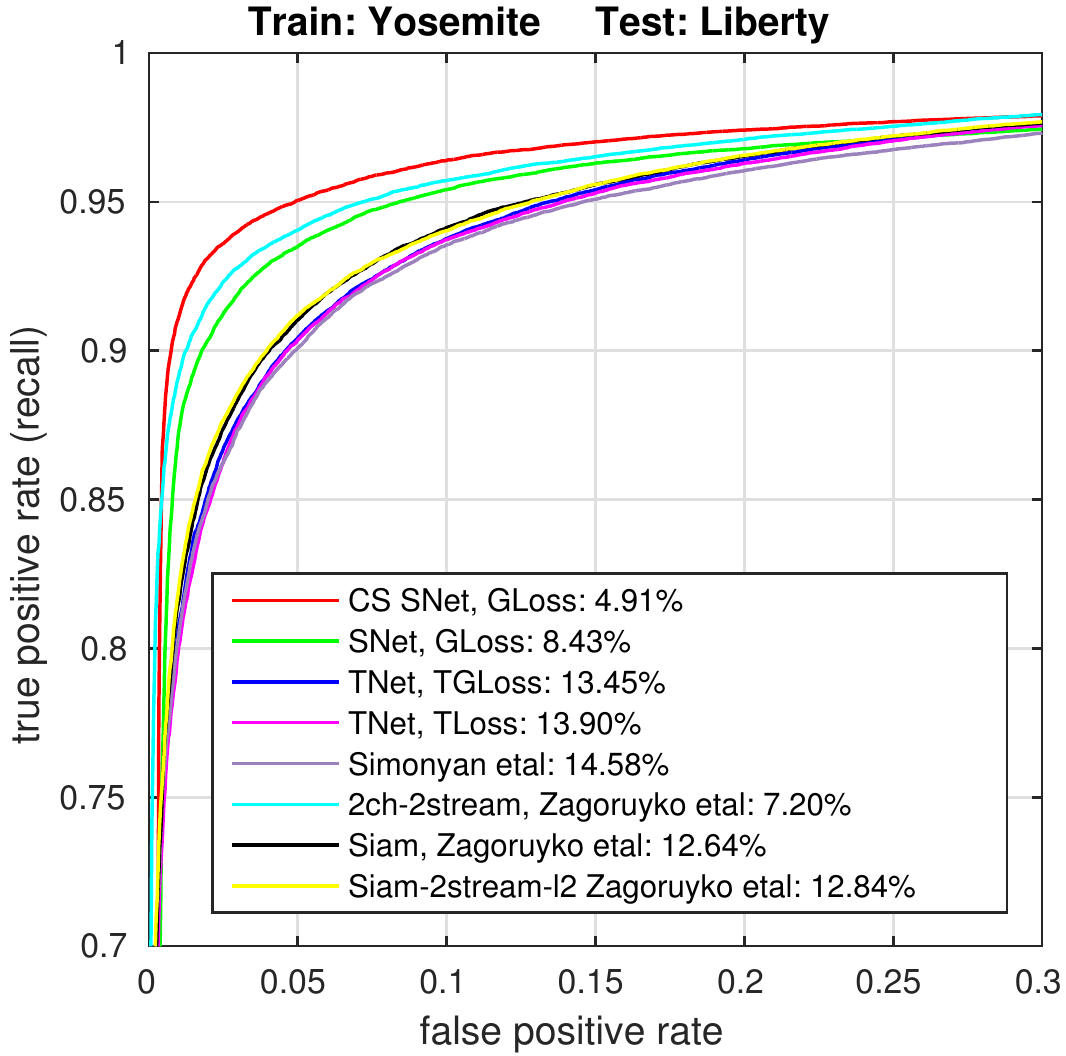} \\
\includegraphics[width=0.27\textwidth]{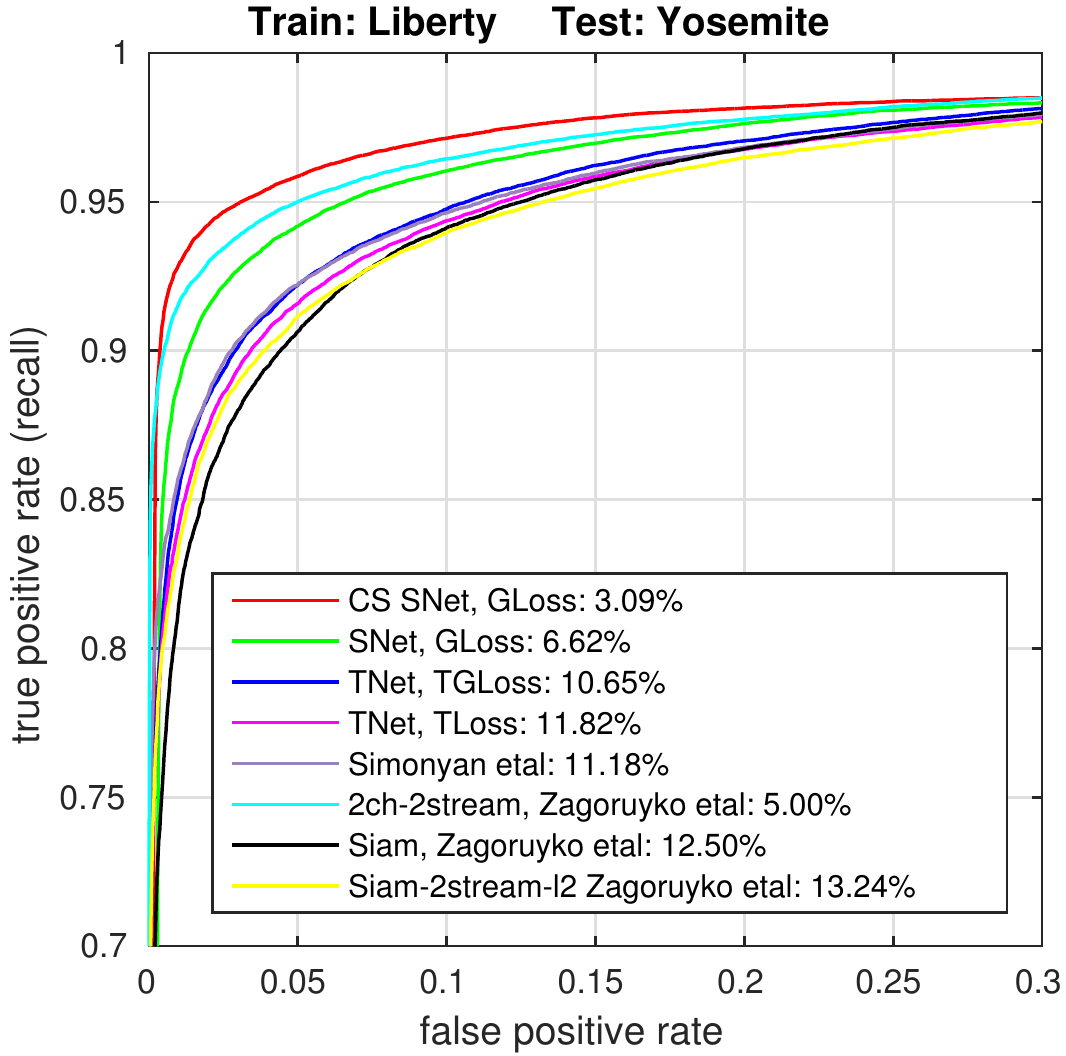} &
\includegraphics[width=0.27\textwidth]{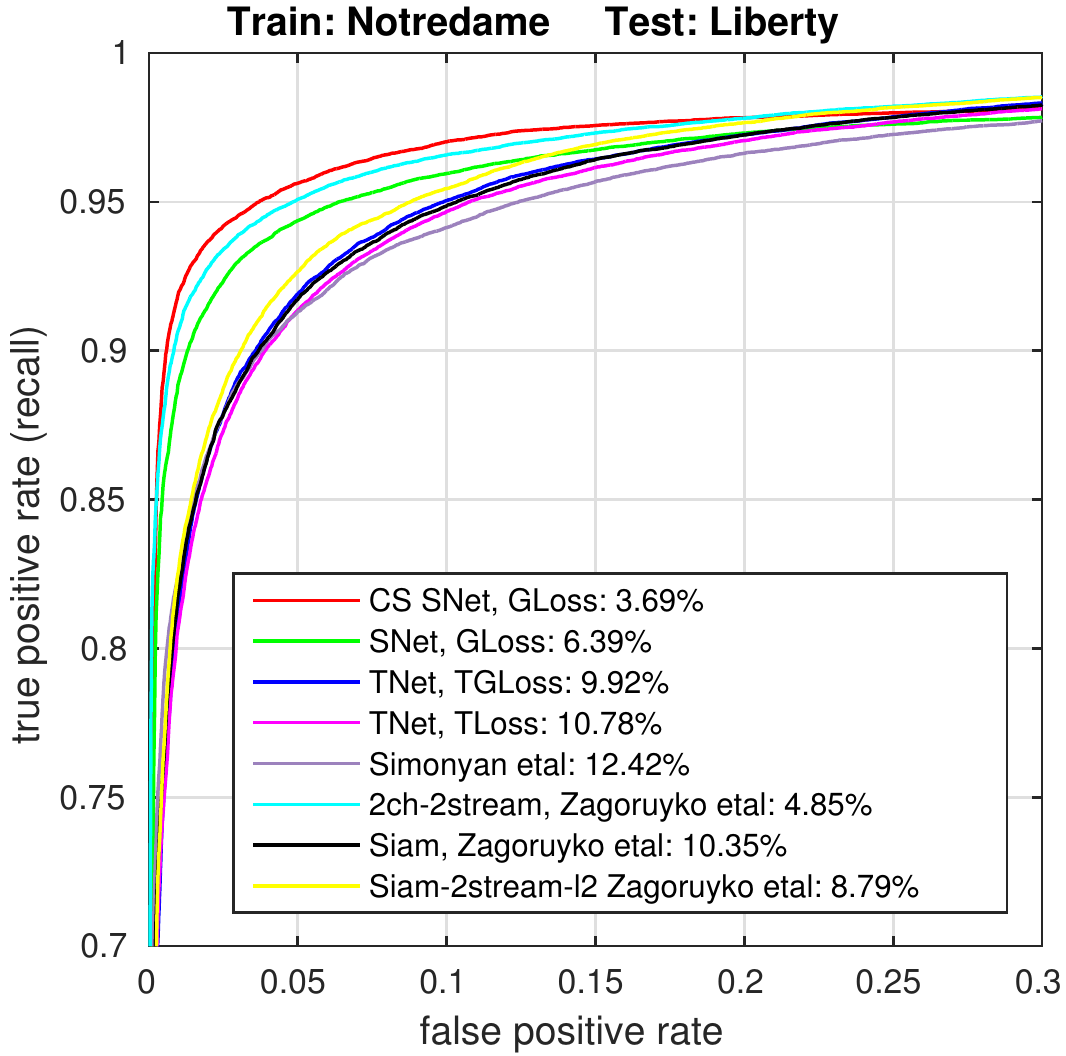} &
\includegraphics[width=0.27\textwidth]{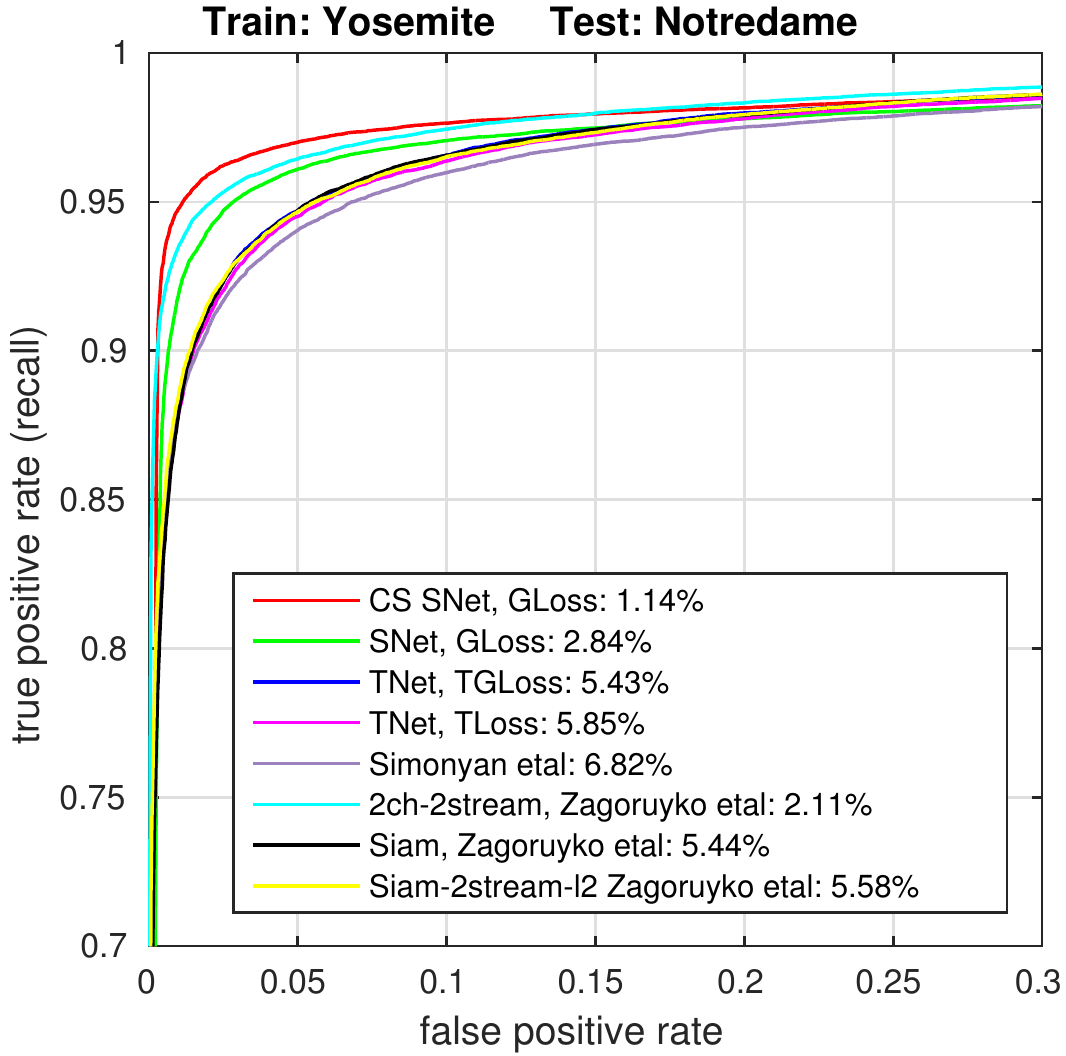} 
\end{tabular}
\end{center}
\caption{ROC curves on the UBC benchmark dataset for our proposed models, and the current state-of-the-art descriptors~\cite{Simonyanpami14,ZagoruykoCVPR15}.  Note that for our models, we use the test set specified in~\cite{UBC_dataset} to compute these curves, and the numbers in the legends represent the FPR95 values.}
\label{fig:rocall}
\end{figure*}

\subsection{Results on UBC Benchmark Dataset}

% 
% \begin{figure}[h]
%  \centering
% \includegraphics[trim = 0.5in 0.3in 0.5in 0.3in, clip,scale=0.5]{filters.pdf}
% \caption{Filters learned in the first convolutional layer of the TNet-TGLoss network}
% \label{fig_filters}
% \end{figure}

Tables~\ref{tbl:table_fpr_pairwise}~and~\ref{tbl:table_fpr_embedding} summarises the performance of the proposed models and compares them with the current state-of-the-art methods for the UBC dataset~\cite{UBC_dataset,brownpami10,snavely2008modeling} using the FPR95 on each of the six combinations of training and testing sets, and the mean over all combinations.  Note that we separate the results in terms of the comparison of descriptors obtained by pairwise similarity methods (Tab.~\ref{tbl:table_fpr_pairwise}) and embedding (Tab.~\ref{tbl:table_fpr_embedding}).  We also show the ROC curves for each of the six combinations of training and testing sets in Fig.~\ref{fig:rocall} for our proposed models, in addition to the current state-of-the-art models~\cite{Simonyanpami14,ZagoruykoCVPR15}.

From the results in Tab.~\ref{tbl:table_fpr_embedding} and Fig.~\ref{fig:rocall}, we observe that the proposed triplet network trained with a combination of the triplet and global losses (\ie, the {TNet-TGLoss}) shows the best result in the field in terms of feature embedding.  The pairwise similarity results in Tab.~\ref{tbl:table_fpr_pairwise}  and Fig.~\ref{fig:rocall} indicate that our centre-surround siamese network trained with global loss (\ie, the {CS SNet, GLoss}) produces a result that is almost half of the previous state-of-the-art result, \ie, the 2ch-2stream~\cite{ZagoruykoCVPR15}.

Similar to~\cite{ZagoruykoCVPR15}, we notice that the siamese networks trained with the pairwise similarity loss achieve better classification performance compared to the feature embeddings produced by the triplet loss, but the dependence of the siamese networks on pairwise inputs is a potential issue during inference in terms of complexity.  For instance, the $\ell_2$ distance norm computation between feature embeddings can be significantly simplified to a cosine distance dot product, since the descriptor norms are equal to 1, while the siamese networks have to measure the similarity using the final fully connected (FC) layer of the network (assuming the features before that FC layer have been pre-computed).
Even though pairwise similarity methods tend to perform better than feature embedding approaches, according to our results and also the results from~\cite{ZagoruykoCVPR15}, it is interesting to notice that our feature embedding model {TNet-TGLoss} performs better than Siam network~\cite{ZagoruykoCVPR15} and the 512d-F(512) network~\cite{Han_cvpr15}, with both representing examples of pairwise similarity methods.

\section{Conclusions}

We have presented new methods for patch matching based on learning using triplet and siamese networks trained with a combination of triplet loss and global loss applied to mini-batches - this is the first time such global loss and triplet network have been applied in patch matching. This new loss overcomes a number of the issues that have previously arisen when using triplet loss, most notably slow or even unreliable convergence.

We argue that the superior results provided by our models are due to the better regularisation provided by the global loss, as shown in Sec.~\ref{sec:toy_problem}.  We have shown our models to
be very effective on the UBC benchmark dataset, delivering state-of-the-art results.

A natural extension of our models is the use of the global loss with the triplet network, but our preliminary results (not shown in this paper)  indicate that this model does not produce better results than the ones in Table~\ref{tbl:table_fpr_embedding}.
We plan to extend this method to other applications, such as pre-training in visual class recognition problems.
\\\\
\textbf{Acknowledgements:} This research was supported by the Australian Research Council through the Centre of Excellence in Robotic Vision, CE140100016, and through Laureate Fellowship FL130100102 to IDR
{\small
\bibliographystyle{ieee}
\bibliography{egbib}
}

\end{document}